# Improving Food Image Recognition with Noisy Vision Transformer

Tonmoy Ghosh, and Edward Sazonov, *Senior Member, IEEE*

*Abstract*— Food image recognition is a challenging task in computer vision due to the high variability and complexity of food images. In this study, we investigate the potential of Noisy Vision Transformers (NoisyViT) for improving food classification performance. By introducing noise into the learning process, NoisyViT reduces task complexity and adjusts the entropy of the system, leading to enhanced model accuracy. We fine-tune NoisyViT on three benchmark datasets: Food2K (2,000 categories, ~1M images), Food-101 (101 categories, ~100K images), and CNFOOD-241 (241 categories, ~190K images). The performance of NoisyViT is evaluated against state-of-the-art food recognition models. Our results demonstrate that NoisyViT achieves Top-1 accuracies of 95%, 99.5%, and 96.6% on Food2K, Food-101, and CNFOOD-241, respectively, significantly outperforming existing approaches. This study underscores the potential of NoisyViT for dietary assessment, nutritional monitoring, and healthcare applications, paving the way for future advancements in vision-based food computing. Code for reproducing NoisyViT for food recognition is available at [NoisyViT_Food](NoisyViT_Food).

*Clinical Relevance*— Accurate food image recognition enhances dietary monitoring, disease management, and personalized healthcare.

## I. INTRODUCTION

Food image recognition, a computer vision task involving the identification and classification of food items, has gained significant attention due to its broad applications in food quality assessment [1], dietary assessment [2], food logging [1], and food intake monitoring [3], [4]. While advancements in computer vision and deep learning have markedly improved the accuracy of food image classification, real-world challenges persist. Factors such as variability in food appearance, occlusions, and the complexity of dining environments continue to pose difficulties.

Before 2010 and the advent of deep learning with large image datasets, traditional machine learning algorithms were the standard approach for food category recognition [5]. In recent years, deep learning has become a dominant force in the machine learning domain, with convolutional neural networks (CNNs) being widely adopted for food recognition tasks [2], [4]. Models such as the Deep Convolutional Neural Network (DCNN) based on the ResNet-50 architecture have demonstrated success in various studies [9]. Similarly, Progressive Region Enhancement Network (PRENet), which combines local and regional features, has been proposed for recognizing food items in the Food2K dataset, which comprises 2,000 food classes [10].

More recently, transformer-based models have shown promising potential for improving food recognition accuracy by capturing long-range dependencies in visual data [8], [9]. Approaches such as the Swin Transformer, Vision Transformer (ViT), and Data-Efficient Image Transformer (DeiT) have been explored, achieving incremental accuracy improvements [13]. For example, a modified Swin Transformer was reported to enhance accuracy in food detection tasks [14]. Another Vision transformer-based study used for real-time food recognition is reported in [15]. The VMamba-based classifier, which incorporates a visual state space model requiring less attention than self-attention, also demonstrated modest improvements in accuracy [16]. High-temperaturE Refinement and Background Suppression network (HERBS) [17] was proposed for food recognition on CNFOOD-241 dataset. While these models treat food recognition primarily as a classification problem, other studies have approached it as an object detection challenge.

In object detection, methods like LOFI (LOng-tailed FIne-Grained Network), which employs an R-CNN-based network, have been proposed for detecting food items [18]. Similarly, YOLOv5 has been applied to food item recognition [19]. Food segmentation methods, such as those leveraging the Segment Anything Model (SAM), have also been explored [20]. However, object detection and segmentation approach often face limitations due to smaller datasets and fewer classes, resulting in lower recognition accuracy compared to classification-based methods.

In this study, we aim to enhance food recognition accuracy on the Food2K, Food-101, and CNFOOD-241 datasets by leveraging the recently introduced Noisy Vision Transformer (NoisyViT) [21] classifier. We fine-tune NoisyViT on the training sets of each dataset using pre-trained weights from Vision Transformer (ViT) trained on ImageNet. The performance of the trained model is then evaluated on the respective test sets. Finally, we compare NoisyViT's food recognition accuracy against existing state-of-the-art models. The remainder of this paper is organized as follows: Section 2 describes the methodology, Section 3 presents the results and discussions, and Section 4 concludes the study.

*The National Institutes of Health supported research reported in this publication through the National Institute of Diabetes and Digestive and Kidney Diseases award number R01DK122473. The content was solely the responsibility of the authors and does not necessarily represent the official views of the National Institutes of Health.

Tonmoy Ghosh is with the Electrical and Computer Engineering Department, University of Alabama, Tuscaloosa, AL 35401 USA (e-mail: tghosh@crimson.ua.edu).
Edward Sazonov is with the Electrical and Computer Engineering Department, University of Alabama, Tuscaloosa, AL 35401 USA (e-mail: esazonov@eng.ua.edu).

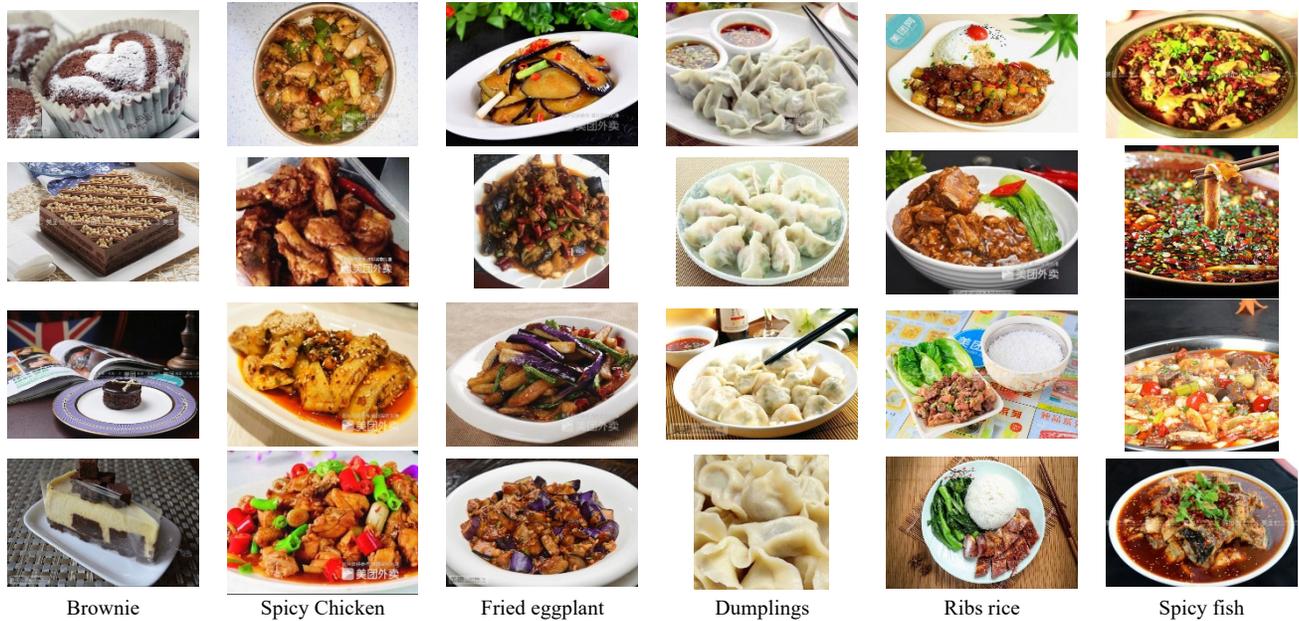

| | | | | | |
|---|---|---|---|---|---|
| Brownie | Spicy Chicken | Fried eggplant | Dumplings | Ribs rice | Spicy fish |

Fig. 1. Sample food images from Food2k dataset

## II. METHODOLOGY

### A. Dataset

Food2K dataset [10] is one of the most comprehensive and challenging benchmarks introduced for food recognition tasks. It contains 1,036,564 color images of food, organized into 2,000 categories, making it significantly larger than existing food recognition datasets in both the number of categories and images—surpassing them by order of magnitude. This establishes Food2K as a highly challenging benchmark for developing advanced models in food visual representation learning. Examples of food categories included in Food2K are illustrated in Fig. 1.

Food2K is structured with 12 super-classes and 26 sub-classes, encompassing a broad range of commonly recognized food types. The 12 super-classes include cereal, vegetables, bread, snacks, soup & porridge, barbecue, egg products, dessert, beans, seafood, fried food, and meat. Despite its extensive coverage, Food2K presents notable class imbalances, with the number of images per category ranging from 153 to 1,999. Additionally, the image dimensions vary between 220 and 597 pixels for both width and height.

To support model development and evaluation, Food2K provides a standardized split of the dataset into training, validation, and testing subsets. Specifically, 620,192 images (59.83%) are allocated for training, 104,513 images (10.08%) for validation, and 311,859 images (30.09%) for testing. This carefully curated dataset, with its scale and diversity, offers a unique opportunity for researchers to develop robust and generalized models.

CNFOOD241 is a large-scale dataset comprising 190,000 images across 241 diverse food categories. The images maintain their original aspect ratios while being standardized to a resolution of 600 × 600 pixels. The dataset is organized into two folders: 'train600x600' and 'val600x600'. The training folder ('train600x600') contains 170,868 images, which were used for model training. The validation folder ('val600x600') includes 20,943 images, which were used for evaluation.

The Food-101 dataset [22] comprises 101,000 food images categorized into 101 distinct food classes, encompassing a diverse selection of dishes from various global cuisines. These categories include widely recognized items such as pizza, sushi, burgers, and salads, among others. Each category contains 1,000 images, which were sourced from multiple platforms, including popular cooking websites and photo-sharing services. The dataset includes a predefined split, with 75,750 images designated for training and 25,250 images for testing, facilitating standardized evaluation across different models.

### B. Data preprocessing and augmentation

The performance of Vision Transformers (ViTs) is highly dependent on augmentation and regularization techniques [23]. To achieve optimal performance, we applied the RandAugment method [24] to augment the training images. Specifically, we employed two random sequential augmentation transformations, setting the magnitude for all transformations to nine. The interpolation mode was set to 'Nearest'. Following augmentation, the training images were randomly cropped and then resized to match the input dimensions of the ViTs. Finally, the images were normalized by subtracting the mean value from each channel and dividing by the standard deviation. For validation and testing images, no augmentation was applied. Instead, they were resized to match the input dimensions of the ViTs and underwent the same normalization process as the training images.

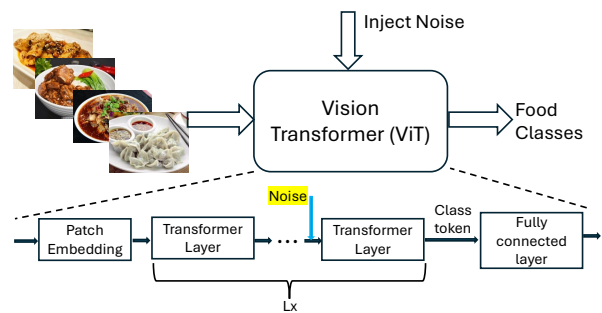

Fig. 2. Food recognition framework using NoisyViT

*C. Noisy Vision Transformer (NoisyViT)*

NoisyViT, introduced in [21], is a universal framework designed for classification tasks. The framework, illustrated in Fig. 2, integrates Vision Transformers (ViTs) with a novel approach that involves injecting noise into a randomly selected layer. This simple operation enables the model to access additional information, reducing task complexity and enhancing performance. Once a specific layer is chosen for noise injection, it remains fixed throughout both training and inference phases.

The findings in [21] demonstrate that noise can effectively alter the entropy of the learning system, leading to improved recognition accuracy. Specifically, noise is introduced into a ViT layer through a linear transformation of the feature matrix, referred to as linear transform noise. This injection in the latent space simplifies the complexity of the task and modifies the system's entropy. The entropy changes, anticipated to be positive, are modeled as an optimization problem. The goal is to determine an optimal quality matrix Q, that maximizes the entropy change while adhering to the constraints imposed by the ViTs. The upper limit of the entropy change induced by positive noise is influenced by the dataset's scale, i.e., the number of data samples. Consequently, larger datasets yield more pronounced benefits from injecting positive noise into deep learning models.

In this study, we utilized the base version of the Vision Transformer (ViT) model [21], which consists of 12 layers and a patch size of 16. The model was initialized with pre-trained weights from ImageNet-1K. Experiments were conducted using two image resolutions: 224 × 224 and 384 × 384. Linear noise was injected into the final layers of the model. For training, we employed a learning rate of 1e-5, the AdamW optimizer, a cosine learning rate scheduler, and label-smoothing-cross-entropy as the loss function. Since we fine-tuned the NoisyViT classifier for food recognition, the model was trained for 30 epochs. The model was fine-tuned and evaluated on a system equipped with a 13th Gen Intel® Core™ i9-13900K CPU, 64GB RAM, and an NVIDIA RTX 6000 Ada Generation GPU. Upon completion of training, the best-performing model was selected for inference on the test dataset, with the results reported in the results and discussion section. To evaluate the performance of the trained models, we used Top-1 and Top-5 accuracy metrics. Additionally, we compared our results with those of state-of-the-art models to assess the effectiveness of our approach.

### III. RESULTS AND DISCUSSION

*A. Performance on Food2K*

Fig 3. illustrates the training and validation loss, as well as accuracy trends over 30 epochs using NoisyViT with a resolution of 384 × 384. The loss values for both training and validation progressively decrease and eventually stabilize. Similarly, training accuracy, along with validation Top-1 and

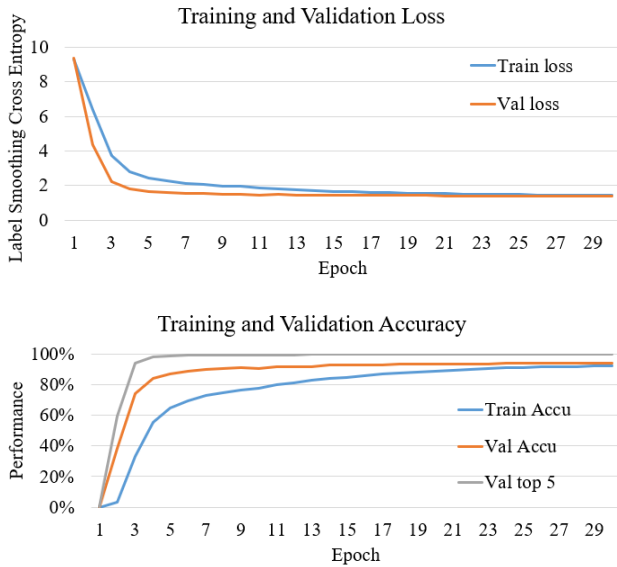

Fig. 3. Training and Validation on Food2K dataset

Top-5 accuracy, increases with more training epochs and eventually plateaus. At the end of training, we achieved 92.2% training accuracy, 94.1% Top-1 validation accuracy, and 99.7% Top-5 validation accuracy. The model with the highest Top-1 validation accuracy was saved and used for inference on the test dataset, which contained 311,859 images across 2,000 classes. On the test set, the model achieved 95% Top-1 accuracy and 99.8% Top-5 accuracy, as reported in Table I (bottom row). NoisyViT with a 384 × 384 resolution consists of 348 million parameters and has an inference time of 6.81ms per image. We also evaluated NoisyViT with a 224 × 224 resolution, and the results are reported in Table I (second row from the bottom). This variant achieved 94.1% Top-1 accuracy and 99.8% Top-5 accuracy, closely matching the performance of the higher-resolution model. However, it has only 86

TABLE I. FOOD RECOGNITION ACCURACY ON FOOD2K DATASET

| Models | Params | Top-1 Acc | Top-5 Acc |
| --- | --- | --- | --- |
| PreNet [10] | NA | 83.75% | 97.33% |
| ViT-B [13] | 86M | 78.41% | 96.33% |
| Swin-B [13] | 88M | 77.58% | 96.17% |
| DeiT-B [13] | 86M | 73.45% | 94.42% |
| R50 + VIT_B_16 [15] | NA | 84.10% | 96.20% |
| **NoisyViT-B 224 X 224** | **86M** | **94.10%** | **99.80%** |
| **NoisyViT-B 384 X 384** | **348M** | **95.00%** | **99.80%** |

NA: not available, B: base model

TABLE II. FOOD RECOGNITION ACCURACY ON FOOD101 DATASET

| Models | Params | Top-1 Acc | Top-5 Acc |
| --- | --- | --- | --- |
| ViT-B [13] | 86M | 88.46% | 98.05% |
| Swin-B [13] | 88M | 92.67% | 98.95% |
| R50 + VIT_B_16 [15] | NA | 91.30% | 99.00% |
| **NoisyViT-B 224 X 224** | **86M** | **99.50%** | **100.00%** |

TABLE III. FOOD RECOGNITION ACCURACY ON CNFOOD-241 DATASET

| Models | Params | Top-1 Acc | Top-5 Acc |
| --- | --- | --- | --- |
| R50 + VIT_B_16 [15] | NA | 83.40% | 95.20% |
| Res-VMamba [16] | 50M | 81.70% | 96.83% |
| HERBS [17] | NA | 82.72% | 97.19% |
| **NoisyViT-B 224 X 224** | **86M** | **96.60%** | **99.90%** |

million parameters and a significantly lower inference time of 2.05ms per image, making it a more efficient choice for food recognition applications.

The performance of NoisyViT on the Food2K dataset is compared with previously published studies, as summarized in Table I. The PreNet model [10], based on ResNet, achieved a Top-1 accuracy of 83.75%. Meanwhile, ViT-B, Swin-B, and DeiT-B [13] reported Top-1 accuracies of 78%, 77%, and 73%, respectively, with Top-5 accuracies ranging between 94% and 97%. In contrast, NoisyViT outperforms these models, improving Top-1 accuracy by at least 10% and Top-5 accuracy by 2.5%. To the best of our knowledge, NoisyViT achieves the highest Top-1 and Top-5 accuracy reported on the large-scale Food2K dataset.

*B. Performance on FOOD-101*

The performance of NoisyViT on the Food-101 dataset is evaluated and compared with previously published studies, as summarized in Table II. NoisyViT achieves a Top-1 accuracy of 99.5% and a Top-5 accuracy of 100% on the dataset. The results are compared with recent models, including ViT-B, Swin-B, and R50+ViT_B_16. NoisyViT demonstrates a 7–11% improvement in Top-1 accuracy and a 1–2% improvement in Top-5 accuracy, highlighting its superior performance over existing methods.

*C. Performance on CNFOOD-241*

The performance of NoisyViT on the CNFood-241 dataset is evaluated and compared with previously published studies, as summarized in Table III. NoisyViT achieves a Top-1 accuracy of 96.6% and a Top-5 accuracy of 99.9% on the dataset. The results are compared with recent models, including Res-VMamba, HERBS, and R50+ViT_B_16. NoisyViT outperforms these approaches, demonstrating a 13–15% improvement in Top-1 accuracy and a 2.5–4% improvement in Top-5 accuracy, establishing its effectiveness in food classification tasks.

## IV. CONCLUSION

In this study, we fine-tuned the NoisyViT classifier for the food recognition task and evaluated its performance on three publicly available benchmark datasets. On the largest Food2K dataset (~1 million images across 2,000 food classes), NoisyViT achieved a Top-1 accuracy of 95%, representing an 11% improvement over the state-of-the-art food recognition models. Similarly, on the Food-101 dataset (~100K images, 101 food classes), NoisyViT achieved a Top-1 accuracy of 99.5%, outperforming the best existing model by 7%. For the CNFOOD-241 dataset (~190K images, 241 food classes), NoisyViT reached a Top-1 accuracy of 96.6%, marking a 13% improvement over prior studies. The significant accuracy gains achieved by NoisyViT demonstrate its effectiveness in food item classification, making it a valuable tool for advancing vision-based food computing applications. This includes food intake monitoring in healthcare, dietary assessment, disease management (e.g., diabetes, hypertension, kidney disease), and personalized healthcare solutions. In future work, we plan to integrate automatic food item recognition into the I2N software [2], enabling real-time monitoring of eating behavior using wearable sensors [25] and cameras [3].